\documentclass[letterpaper,twocolumn,10pt]{article}
\usepackage{usenix2019_v3}

\usepackage{tikz}
\usepackage{amsmath}

\usepackage{multirow}
\usepackage{booktabs}
\usepackage{graphicx}
\usepackage{amsmath}
\usepackage{amssymb}

\usepackage{filecontents}

\usepackage{appendix}

\begin{document}

\date{}

\title{\Large \bf Deciphering Textual Authenticity: A Generalized Strategy through the Lens of Large Language Semantics for Detecting Human vs. Machine-Generated Text}


\author{
{\rm Mazal Bethany\textsuperscript{1,2}, Brandon Wherry\textsuperscript{1,2}, Emet Bethany\textsuperscript{1,2},}\\
{\rm Nishant Vishwamitra\textsuperscript{1}, Anthony Rios\textsuperscript{1}, Peyman Najafirad\textsuperscript{1,2}\thanks{Corresponding Author}}\\
\textsuperscript{1}The University of Texas at San Antonio\\
\textsuperscript{2}Secure AI and Autonomy Lab
}

\maketitle



\begin{abstract}
\label{abstract}

With the recent proliferation of Large Language Models (LLMs), there has been an increasing demand for tools to detect machine-generated text. The effective detection of machine-generated text face two pertinent problems: First, they are severely limited in generalizing against real-world scenarios, where machine-generated text is produced by a variety of generators and spans diverse domains. Second, existing detection methodologies treat texts produced by LLMs through a restrictive binary classification lens, neglecting the nuanced diversity of artifacts generated by different LLMs, each of which exhibits distinctive stylistic and structural elements. %
In this work, we undertake a systematic study on the detection of machine-generated text in real-world scenarios. 
We first study the effectiveness of state-of-the-art approaches and find that they are severely limited against text produced by diverse generators and domains in the real world. Furthermore, t-SNE visualizations of the embeddings from a pretrained LLM's encoder show that they cannot reliably distinguish between human and machine-generated text.
Based on our findings, we introduce a novel system, T5LLMCipher, for detecting machine-generated text using a pretrained T5 encoder combined with LLM embedding sub-clustering to address the text produced by diverse generators and domains in the real world.
We evaluate our approach across 9 machine-generated text systems and 9 domains and find that our approach provides state-of-the-art generalization ability, with an average increase in F1 score on machine-generated text of 11.9\% on unseen generators and domains compared to the top performing supervised learning approaches and correctly attributes the generator of text with an accuracy of 93.6\%.  
We make the code for our proposed approach publicly available at \url{https://github.com/SecureAIAutonomyLab/LLM-Cipher}
\end{abstract}

\section{Introduction}
\label{introduction}
In the recent years, we have witnessed the explosive adoption of Large Language Models (LLMs), ushering in a new era of natural language processing. LLMs, such as ChatGPT~\cite{chatgpt2023}, have demonstrated remarkable proficiency in generating human-like text, comprehending intricate nuances, and even performing tasks that traditionally required human intuition. Despite these successes, there has been increasing concern around these publicly available models being used in problematic applications such as to spread fake news~\cite{bhat2020effectively}. In addition, there's evidence showcasing the ability to adapt and fine-tune LLMs for specific purposes quickly. Notably, Grover~\cite{zellers2019defending}, which is based on the GPT-2~\cite{openaiGPT2} architecture, was introduced just months after the release of GPT-2. In light of the advancements and capabilities of LLMs, there has been a burgeoning interest in discerning machine-generated text from human-generated text~\cite{crothers2023machine}.

Although strides have been made in identifying machine-generated text in recent times~\cite{mitchell2023detect}, there has been little research in working towards methods that can effectively distinguish human and machine-generated text in real-world scenarios by malicious actors who have access to wide-ranging resources. 
Detecting machine-generated text in real-world scenarios faces two pertinent problems. \textit{First}, given the escalating proliferation and sophistication of LLMs, adversaries can utilize a wide variety of LLMs through publicly available models~\cite{huggingface} or even use one of the many publicly available APIs~\cite{kafkai,aiwriter,articleforge} to generate machine-generated text. In addition to utilizing a variety of models for text generation, adversaries can produce content in a wide range of domains~\cite{shridhar2023distilling}, which presents a formidable challenge for current detection mechanisms. 
While existing detectors perform effectively on text generated from known generators or domains, they cannot generalize well on machine-generated text from diverse generators or domains~\cite{pu2023deepfake}. \textit{Second}, existing approaches for detecting machine-generated text consider these diverse generators as belonging to a single distribution. However, each text generator creates text that exhibits distinctive stylistic and structural elements~\cite{he2023mgtbench}. Considering the detection of machine-generated text as a binary problem oversimplifies the nuanced differences between generators and potentially reduces the effectiveness of detectors on unseen generators.  

In this work, we take the first step towards developing a generalized strategy for detecting human vs. machine-generated text across diverse generators and domains. Given the challenge of texts produced by various generators across multiple domains, we carried out a study to detect machine-generated content in real-world settings, taking into account an array of generators and domains. We begin by evaluating the performance of current state-of-the-art methods~\cite{pu2023deepfake}. Our findings suggest that these techniques face severe challenges in pinpointing text from diverse generators and across different domains with the top performing model showing an average F1 score of just 53.7\% on this diverse data. 
Upon examining the distribution of embeddings from a pretrained LLM's encoder from a t-SNE visualization, we discover that there is potential in utilizing this embedding for differentiating between human and machine-generated texts.

Based on this observation, we propose to build a system designed to detect machine-generated text in real-world scenarios. We broadly address the following research questions: 1) Can we show how previous methods fail to generalize? 2) Can the encoders from LLMs be used to underpin an approach for generalized machine-generated text detection? 3) Are there fingerprints that can be detected to distinguish between multiple text generators?
Guided by these research questions, we develop our system to detect machine-generated text. We present our system, T5LLMCipher, which uses a pretrained LLM encoder to extract the embeddings of diverse texts and improves generalization to in-the-wild machine-generated text by framing it as a multi-class classification problem.
Investigating 9 machine text generators and 9 text domains, we find that our top-performing detector is able to correctly attribute the generator of a text with an accuracy of 93.6\% and outperforms existing baselines by 19.6\% for the task of detecting human vs machine-generated text on unseen domains and generators of machine-generated text, showing that these generators have specific fingerprints that can be detected. We further show in these experiments that multiclass classifiers may generalize better than binary classifiers for the binary classification task of detecting human vs machine-generated text in-the-wild. Then, we investigate classification architectures towards the analysis of the embeddings from the LLM encoders and show that classifiers built on these embeddings can be adversarially robust and provide state-of-the-art generalization to unseen generators and domains for machine-generated text detection.

Our work makes the following contributions:
\begin{itemize}
    \item \textbf{New Findings.} We study the limitations of  state-of-the-art methods in detecting machine-generated text created in real-world scenarios. Furthermore, we show that the embeddings from LLM encoders can be leveraged to robustly discern human-generated from machine-generated text. Our findings shed light on the severe limitation of existing methods on real-world machine generated text, and paves the way for developing methods that can detect machine-generated text in a generalized manner.
    \item \textbf{New Detection System.}
    We present a new machine-generated text detection system, T5LLMCipher, to robustly detect text as being machine generated using the embeddings from LLM encoders.
    \item \textbf{Comprehensive Evaluation.} We demonstrate that classifiers trained for the multi-class classification task of generator attribution show superior generalization and adversarial robustness to binary supervised classifiers in detecting in-the-wild machine-generated text.
\end{itemize}

\section{Background}
\label{background}

As the digital landscape evolves, ensuring security on social media platforms has become a paramount concern. One salient issue is the spread of false information. Platforms like X (formerly Twitter)~\cite{twitter} have attempted to address this by labeling misleading content. However, these interventions often lack uniformity, leaving considerable amounts of deceptive content unflagged  ~\cite{paudel2023lambretta}. Systems like LAMBRETTA have emerged as a response, offering automated solutions that outpace traditional methods by identifying and flagging erroneous information ~\cite{paudel2023lambretta}. Beyond misinformation, a subtler but equally concerning issue is the proliferation of machine-generated profiles. LinkedIn, a prime example, is combating the rise of such fake profiles, as they pose risks ranging from privacy breaches to phishing attempts. Techniques like the Section and Subsection Tag Embedding (SSTE) have been developed to differentiate between legitimate and artificial profiles, even those generated by sophisticated language models ~\cite{ayoobi2023looming}. The weaponization of social media, from manipulating public opinion to outright media warfare, only intensifies the need for robust security measures ~\cite{chen2022social}. Amid these challenges, the detection of "toxic comments" also remains a focal point, with tools like the regression vector voting classifier (RVVC) aiming to combat online hostility ~\cite{rupapara2021impact}. As these various threats underscore, there's a pressing need for advanced detection mechanisms. Leading into this context, the challenge of discerning machine-generated text from human-authored content has emerged as a crucial problem, necessitating the development of robust detection methodologies.

In the realm of detecting machine-generated text, the literature has predominantly converged around three overarching strategies ~\cite{crothers2023machine}. Many of the first approaches for detecting machine-generated text were feature-based approaches, which emphasize a myriad of characteristics, encompassing word frequency, a spectrum of statistical features, fluency markers, and the consistency of linguistic attributes. More recently, there has been increasing focus on approaches to zero-shot methodologies, which intriguingly deploy the generative models themselves as tools to pinpoint machine-crafted content. Finally, there exists a set of techniques that engage in the fine-tuning of language models specifically for the detection of text generated by machines.

One of the subsets of statistical features leveraged in detecting machine-generated text revolves around term frequencies within text samples. Text authored by humans typically aligns with Zipf’s Law, where the most frequent word in human-written text has roughly twice the frequency of the second most frequent word, and nearly three times that of the third most frequent word and so on ~\cite{nguyen2017identifying}. Another feature-based method for detecting machine-generated text focuses on clarity and coherence. Earlier models like GPT-2 have shown that in longer outputs, there is a higher likelihood that inconsistencies, redundancy, or logical errors will emerge ~\cite{see2019massively}. Some works build classifiers on bag-of-words features and TF-IDF ~\cite{fagni2021tweepfake} to analyze shorter text sequences. Other works such as Fr{\"o}hling et al. ~\cite{frohling2021feature} combine several features, including lack of syntactic and lexical diversity, repetitiveness, and extract linguistic features such as named entities and part-of-speech distributions to build a neural network classifier on this information. However, a notable limitation of these statistical and feature-based methods is their potential vulnerability to more advanced language models, which continually improve in generating human-like text patterns and linguistic diversity.


Zero-shot methods for detecting machine-generated text have garnered attention due to the lack of need to train models on large datasets. Works such as those by Gehrmann et al. ~\cite{gehrmann2019gltr} find that human-written texts exhibit a broader range of word choices, whereas machine-generated texts tend to be more constrained and predictable and calculate entropy on text sequences to help distinguish machine-generated text from human-generated text. Other works used simple log-probability thresholding, where LLM token probabilities are averaged and a threshold is determined ~\cite{solaiman2019release}. Recent works like DetectGPT ~\cite{mitchell2023detect} take this idea further by introducing the observation that the curvature of a model's log probability function is typically more negative for machine-generated text compared to human-written text. Building on this insight, their algorithm leverages the hypothesis by approximating the log probability function's Hessian, and thresholding based off of the difference between perturbed samples and the input sample. However, these methods require access to the token probabilities of the models they are trying to detect.

Another class of methods is fine-tuning methods which utilize pretrained language models to identify machine-generated text. One detection method leverages energy-based models alongside a RoBERTa model ~\cite{liu2019roberta}, which were fine-tuned on datasets of machine-generated text ~\cite{bakhtin2021residual}. More recent works including GPT-Sentinel also leverage a pretrained T5 model ~\cite{raffel2020exploring} and use a multi-layer-perceptron module to classify the hidden state vector of input, to directly encode the task as a sequence-to-sequence (seq-to-seq) problem ~\cite{chen2023gpt}. Among this class of works, it is consistently demonstrated that the performance of fine-tuned models is heavily influenced by the specific models and datasets on which they are trained. There is a prevailing observation that these models struggle to generalize effectively when encountering generative models not present in their fine-tuning datasets. Additionally, for optimal performance, it is essential that these models are exposed to a diverse array of topics and writing styles during their training phase. Furthermore, the generation of these datasets may also prove to be expensive if attempting to include a greater and greater number of LLMs within the dataset to aid in generalization.

Overall, while these approaches have been shown to generalize well when applied to different LLMs, but have not necessarily shown good generalization \emph{between} different LLMs and text domains. Given the growing number and complexity of LLMs, generalizable systems for detecting machine-generated text are of upmost importance.

\section{Threat Model and Problem Scope}
\label{ThreatModelAndProblemScope}
\subsection{LLM Text Generation Objective}

The core objective during the training of autoregressive language models is captured by the equation:

\begin{equation}
\label{eq:llm}
L(\theta) = -\log \prod_{i=1}^{n} f_{\theta}(x_i | x_1, \dots, x_{i-1})
\end{equation}

This equation portrays the negative log-likelihood of a model with parameters $\theta$, where the objective is to predict each token in a sequence based on its preceding tokens. Minimizing this likelihood value signifies that the model assigns higher probabilities to observed sequences, indicating a better fit to the training data.

However, as straightforward as this equation may appear, the ultimate behavior and artifacts of an LLM are influenced by several factors. The distribution and nature of the training data shape the model's behavior. For instance, an LLM trained on a broad, multi-domain dataset might demonstrate extensive generalization capabilities. In contrast, a model trained on niche data could display a more narrow understanding. Patterns present in the training data may manifest as artifacts in the machine-generated text. The model architecture further complicates how Equation~\ref{eq:llm} is learned. Choices surrounding the number of layers, attention mechanisms, and other hyperparameters can lead to differences in the artifacts present in the text outputs.

\subsection{Threat Model}
In the context of machine-generated text detection, we are introduced to two main parties at the heart of our investigation. The first party is comprised of adversaries that produce or disseminate machine-generated text. These adversaries may leverage publicly available models, fine-tune existing models, use existing machine-text API, or even train their own LLM. One of the goals of the adversary is to evade detection of their use of machine-generated text. To evade detection, adversaries may use state-of-the-art LLMs to closely mimic the distribution of human texts or use adversarial attacks to perturb machine texts specifically to avoid detection by automated detectors. LLMs can craft texts that are increasingly indistinguishable from human-generated texts, presenting a complex challenge for detection systems. The second party is made of those invested in distinguishing between human and machine-generated text. A significant constraint faced by these stakeholders is their lack of prior knowledge about which specific LLM has been employed to produce any given text or the domains in which they may generate texts for. The primary objective for this party is to detect and flag content that is solely the product of machine generation.

Our study is anchored in scenarios where language models craft entire paragraphs, setting aside considerations of adversarial prompting ~\cite{maus2023adversarial, zhu2023promptbench} or paraphrasing-based attacks ~\cite{sadasivan2023can} for future exploration. Additionally, we do not delve into detecting texts that were initially machine-generated but later edited by humans. Such texts, with their blend of human and machine nuances, introduce complexities in detection that lie outside of the scope of this work. Instead, our primary aim is centered on the pinpoint identification of text that is solely the product of machine generation.

\section{Data}
\label{Data}
This work leverages diverse datasets collected by Wang et al. ~\cite{wang2023m4} in the M4 dataset to construct a robust detector for machine-generated text. We focus on the English corpora they released so that we can compare our approach against other in-the-wild machine-generated text datasets. We analyze five sources of human-generated data and five LLMs used to generate machine-generated text from the M4 dataset~\cite{wang2023m4}, for a total of 126,950 text samples. These LLMs are ChatGPT (gpt3.5-turbo) ~\cite{chatgptdoc}, davinci-003 ~\cite{chatgptdoc}, Cohere ~\cite{cohere}, Dolly-v2 ~\cite{dollyv2}, and BLOOMz ~\cite{BLOOMz2022}. Additional discussions on dataset assumptions and experiments that assess the quality of the machine-generated texts in the are detailed in Appendix \ref{sec:appendix_quality}. This appendix presents results from an automatic evaluation using Universal Sentence Encoder~\cite{cer2018universal}, alongside human evaluations that assess the stylistic and semantic similarity between the human and machine-generated texts. The human text sources and the corresponding text generation strategies are described below, with further prompt details provided in Appendix \ref{sec:appendix_prompts}:

\begin{itemize}
    \item \textbf{Wikipedia:} Comprises 3,000 articles from Wikipedia, each exceeding 1,000 characters. LLM's are given titles as prompts for generating new articles.
    \item \textbf{Reddit ELI5:} A collection of 3,000 preprocessed question and response pairs from subreddits r/explainlikeimfive, r/askscience, and r/AskHistorians. Questions are used as prompts for the LLM to generate an answer.
    \item \textbf{WikiHow:} Consists of 3,000 articles from WikiHow, each with a title, headline, and text. Articles are generated by LLMs using titles and headlines from the human data as prompts.
    \item \textbf{PeerRead Reviews:} The dataset includes title, abstract, and multiple human reviews of 586 papers from top-tier conferences in NLP and machine learning. Four different prompts to the LLM are used to create machine-generated peer reviews.
    \item \textbf{Arxiv Abstract:} This dataset includes 3,000 abstracts from Arxiv, each with a minimum of 1,000 characters. Abstracts are generated by LLMs using titles from the human data as prompts.
\end{itemize}

For our experiments on in-the-wild data, we use the in-the-wild data released by Pu et al. ~\cite{pu2023deepfake}. Each of these datasets of machine-generated text also contain an equal number of human-generated text from which the machine-generated text was spawned. The details of this data is presented below:

\begin{itemize}
    \item \textbf{AI-Writer:} 1,000 real news articles were scraped from 20 popular news websites, sampled from the RealNews ~\cite{zellers2019defending} dataset. Using the titles from these real articles, 1,000 synthetic articles were generated through the AI-Writer ~\cite{aiwriter} service, which employs custom Transformer-based LMs.
    \item \textbf{ArticleForge:} 1,000 real news articles were collected, and using a set of keywords from these, 1,000 synthetic articles were generated through the ArticleForge ~\cite{articleforge} service, which utilized fine-tuned versions of GPT-2, BERT, and T5 for text generation.
    \item \textbf{Kafkai:} 1,000 real articles from 10 different categories were used as priming text to generate 1,000 synthetic articles through the Kafkai ~\cite{kafkai} service, which used fine-tuned models from OpenAI, including GPT-2.
    \item \textbf{RedditBot:} 887 real comments with a length greater than 192 tokens were randomly sampled from forum threads on Reddit, matched with 887 machine-generated comments generated by a GPT-3 powered bot under the username /u/thegentlemetre on /r/AskReddit.
\end{itemize}


\section{Motivation}
\label{motivation}

\begin{table}[b]
\centering
\resizebox{\columnwidth}{!}{%
\begin{tabular}{@{}rrrrrrrr@{}} 
\toprule
\multirow{2}{*}{Generator} & \multirow{2}{*}{Dataset} & \multicolumn{4}{c}{Detectors} \\ \cmidrule(l){3-6} 
                           &                          & BERT-D & GLTR-BERT & GLTR-GPT2 & RoBERTa-D \\ \midrule 
\multirow{5}{*}{ChatGPT}   & arxiv                    & 7.6 & 66.7    & 11.3    & 48.4    \\ 
                           & peeread                  & 0.0      & 48.1    & 11.3    & 34.5    \\ 
                           & reddit                   & 1.1 & 90.0    & 59.2    & 54.8    \\ 
                           & wikihow                  & 0.4 & 69.0    & 40.0    & 20.0    \\ 
                           & wikipedia                & 0.0      & 77.8    & 50.4    & 59.9    \\ 
\midrule
\multirow{5}{*}{Dolly}     & arxiv                    & 8.2 & 22.6    & 2.0     & 57.8    \\ 
                           & peeread                  & 29.4 & 21.2    & 0.0     & 56.0    \\ 
                           & reddit                   & 11.3 & 76.5    & 3.9     & 67.9    \\ 
                           & wikihow                  & 14.3 & 31.1    & 7.7     & 52.6    \\ 
                           & wikipedia                & 2.0  & 59.4    & 3.9     & 56.3    \\ 
\bottomrule
\end{tabular}%
}
\caption{F1 score with respect to machine-generated text of existing detectors on ChatGPT and Dolly generations across five text domains.}
\label{tab:motivation}
\end{table}

Numerous studies on machine-generated text detection employ a variety of domains and generators for testing purposes \cite{kushnareva2021artificial, aich2022demystifying}. Despite this diversity in evaluation, we demonstrate that many of the models developed and validated by these studies exhibit a failure to generalize effectively across domains and generators. These models, while perhaps adept at handling the specific types of machine-generated text present in their respective training environments, falter when applied to external datasets, underscoring a significant limitation in their practical utility and application.

We conduct a preliminary experiment to show that existing state-of-the-art methods for detecting machine-generated text fail to generalize to unseen data. To accomplish this, we test four of the models presented in the work of Pu et al. ~\cite{pu2023deepfake}: BERT-Defense, GLTR-BERT, GLTR-GPT2, and RoBERTa-Defense. GLTR-BERT and GLTR-GPT2 exploit the insight that synthetic text often contains tokens that are deemed highly probable by language models. GLTR extracts features based on the token probability distributions, such as the number of tokens in the Top-10, Top-100, and Top-1000 ranks. A logistic regression classifier then processes these extracted features. There are two GLTR variants: GLTR-BERT, which utilizes BERT as its backend LM, and GLTR-GPT2, which uses GPT2-XL. The training setup used WebText for real articles and GPT2-XL for synthetic ones, with a training set size of 4,000 articles per class. BERT-Defense is a binary classifier that builds upon the pre-trained BERT-Large LM by adding a classification layer. The defense is trained and tested using the WebText dataset for real articles and GPT2-Large for the synthetic ones, with a training set size of 10,000 articles for each class. RoBERTa-Defense is inspired by BERT-Defense but is built on the pre-trained RoBERTa with the addition of a classification layer. The RoBERTa-Defense model training setup is on the RealNews dataset for genuine articles and GROVER for synthetic ones, which comprises of 5,000 articles for each class. 

We evaluate these models on a recently released dataset of machine-generated text~\cite{wang2023m4}, and show the F1 score of these detectors with respect to the machine-generated text. We test on the Arxiv, PeerRead, Reddit, Wikihow, and Wikipedia domains with the ChatGPT and Dolly generators. Further details about this dataset can be found in Section~\ref{Data}. The results of this experiment can be found in Table~\ref{tab:motivation}. These experiments reveal significant variations in the detection performance when state-of-the-art detectors are evaluated against unseen generators and domains. While the GLTR-BERT model shows good performance on the Reddit domain, this performance does not generalize well overall with an average F1 score of just 56.3. This inconsistency in performance is particularly detrimental in the context of security applications where detectors play a pivotal role in identifying and mitigating malicious activities. In scenarios where security is paramount, the ability of detectors to accurately identify machine-generated content from unknown generators and domains is vital.

\begin{table}[]
\centering
\resizebox{\columnwidth}{!}{%
\begin{tabular}{@{}rrrrrrrr@{}} 
\toprule
\multirow{2}{*}{Generator} & \multirow{2}{*}{Dataset} & \multicolumn{4}{c}{Detectors} \\ \cmidrule(l){3-6} 
                           &                          & BERT-D & GLTR-BERT & GLTR-GPT2 & RoBERTa-D \\ \midrule 
\multirow{5}{*}{ChatGPT}   & arxiv                    & +21.4   & -79.4    & -100    & -74.2    \\ 
                           & peeread                  & NA    & -83.3    & -100    & -90.0    \\ 
                           & reddit                   & NA   & -28.4    & -100    & -86.3    \\ 
                           & wikihow                  & NA    & -37.3    & -100    & -76.7    \\ 
                           & wikipedia                & NA    & -10.2    & -100    & -79.4    \\ 
\midrule
\multirow{5}{*}{Dolly}     & arxiv                    & +27.3    & -77.8    & -100    & -62.4    \\ 
                           & peeread                  & -4.9   & -87.5    & NA    & -71.9    \\ 
                           & reddit                   & +23.5   & -52.8    & -100    & -63.8    \\ 
                           & wikihow                  & +10.3   & -66.7    & -100    & -71.4    \\ 
                           & wikipedia                & -0.0   & -38.9    & -100    & -61.5    \\ 
\bottomrule
\end{tabular}%
}
\caption{$\Delta$R of existing detectors against black-box adversarial attacks. "+" indicates an improvement in performance, "-" indicates a deterioration in performance. "NA" values indicate that the baseline recall was 0.}
\label{tab:motivation_ER}
\end{table}

\begin{figure*}[]
    \centering
    \includegraphics[width=1\linewidth]{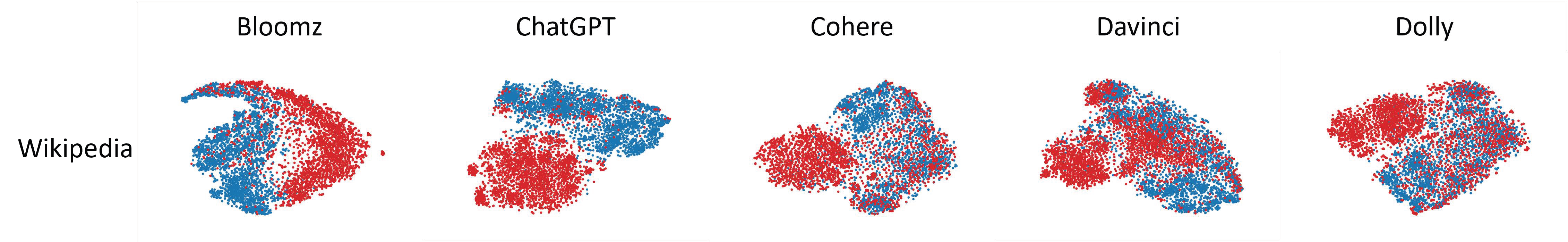}
    \caption{t-SNE visualization of T5 embeddings illustrating the distribution of Human (blue), and Machine (red) generated text. While there is some degree of separation, considerable overlap is still evident.}
    \label{fig:t-SNE_motivation_1}
\end{figure*}

\begin{figure*}[]
    \centering
    \includegraphics[width=1\linewidth]{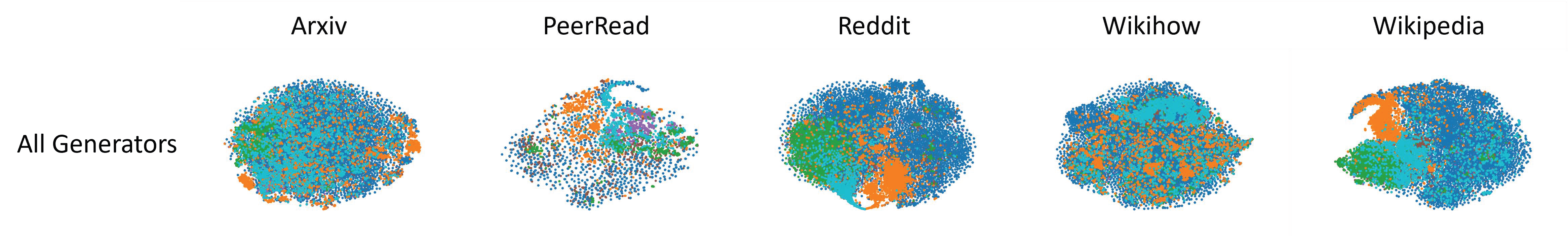}
    \caption{t-SNE visualization of T5 embeddings illustrating the distribution of Human (blue), Bloomz (orange), ChatGPT (green), Cohere (teal), Davinci (purple), and Dolly (brown) generated texts across different domains. It is even more difficult to distinguish the source generator in the multiclass setting.}
    \label{fig:t-SNE_motivation_2}
\end{figure*}

To further show the vulnerability of existing methods, we evaluate the same detectors except for one against a recently introduced adversarial attack, called DFTFooler~\cite{pu2023deepfake}, where we generate adversarial samples designed to evade model detection. Following the evaluation strategy of DFTFooler, we calculate the percentage change in recall, $\Delta$Recall ($\Delta$R). We use this metric to evaluate model resilience to adversarial attacks because it quantifies the change in how well a model correctly identifies the machine-generated text under these attacks. A significant drop in recall indicates the model's vulnerability, as it fails to identify many adversarially altered instances correctly. The results presented in Table~\ref{tab:motivation_ER} underscore the vulnerabilities of current text detection methods against adversarial attacks. Examining the $\Delta$R for both ChatGPT and Dolly generators, we observe a notable trend: GLTR-GPT2 consistently yielded a 100\% drop in recall, highlighting its susceptibility to these attacks irrespective of the text domain. Furthermore, while the performance of detectors such as GLTR-BERT and RoBERTa-D varied based on the dataset, several instances showcased deterioration in performance, ranging from 70-90\%. While the ($\Delta$R) on BERT-D shows positive values, meaning the adversarial attack actually increased the detection performance on adversarially perturbed samples, the baseline recall was already so low that the perturbations flipped the decisions on some of the originally incorrect predictions to being correct predictions. Such high drops in recall by the models accentuate the pressing need for enhancing the robustness of current detection methods, as even these advanced models demonstrate significant vulnerabilities in the face of adversarial attacks.

To underscore a potential method that could be used to robustly identify machine-generated text, we demonstrate that machine-generated text possesses unique characteristics or `fingerprints' that can be identified and analyzed through the encoder of an LLM. For this experiment, we use text embeddings generated using the Google Flan-t5-xl implementation available on Huggingface ~\cite{flan-t5-xl}. This model is known for its effectiveness in capturing the semantic and syntactic nuances of text, making it a suitable choice for generating embeddings that can reveal the subtle differences between human and machine-generated content. Further details regarding the data utilized in this experiment are provided in Section~\ref{Data}. At the same time, insights into the specific parameters and configuration of the T5 model employed are documented in Section~\ref{experiments}. We present the t-SNE of these embeddings to visualize a starting point for our proposed method of using LLM encoders to discern machine-generated text. t-SNE is a non-linear dimensionality reduction technique adept at embedding high-dimensional data for visualization in a two-dimensional space. It calculates the pairwise similarities between data points in the high-dimensional space and endeavors to find a low-dimensional representation that upholds these similarities ~\cite{van2008visualizing}. Importantly, because t-SNE operates as an unsupervised algorithm, it doesn't rely on the true labels of the data. As a result, observing any degree of clustering or separation between classes in the t-SNE visualization is particularly encouraging.

In Figure~\ref{fig:t-SNE_motivation_1}, we present the t-SNE visualization of text embeddings derived from wikipedia text comprising both machine and human-generated text, as detailed in the study by Wang et al. ~\cite{wang2023m4}. 
The preliminary results from the t-SNE experiment are encouraging, showing some separation between human and machine-generated text in the reduced-dimension space created by t-SNE. While there is considerable overlap between the human and machine-generated texts, such initial findings warrant further exploration and analysis into these embeddings. Given that this embedding was never trained to discern human text from machine text, it shows potential to be developed into a robust classifier. The observed separation provides a foundation on which we can apply techniques like contrastive learning or deep neural networks to further refine the distinction between machine-generated and human-generated texts. In the subsequent Figure~\ref{fig:t-SNE_motivation_2}, we delve deeper, showcasing the t-SNE visualizations encompassing six distinct classes: human text and five different machine text generators. Here, the interplay of the classes paints a more complex picture. 
Figure~\ref{fig:t-SNE_motivation_2} shows the increased complexity in distinguishing human-generated text from a diverse array of machine-generated sources, as compared to differentiating human text from a singular machine source. Given that there is some marginal difference in the distribution of the embeddings between the different generators even in this unsupervised visualization suggests that, with further refinement and training, the LLM embedding could be harnessed to attribute the source generator of a specific text piece. This insight broadens the scope of our research, indicating that not only can we detect machine-generated content, but we might also be able to pinpoint its origin.

\section{Approach}
\label{approach}
\begin{figure*}[ht]
    \centering
    \includegraphics[width=0.8\linewidth]{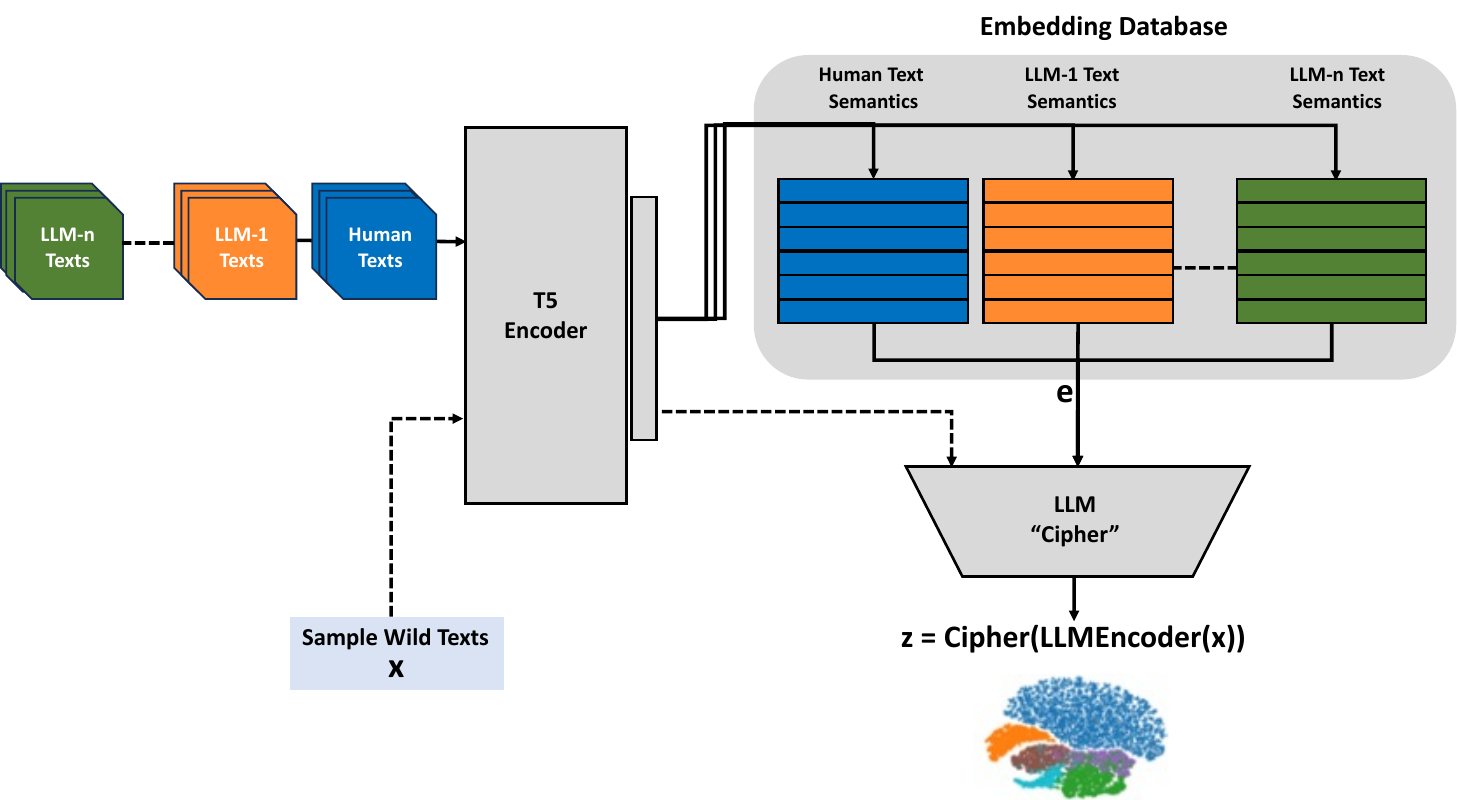}
    \caption{System Architecture of T5LLMCipher for distinguishing between human and machine-generated texts. The architecture consists of an LLM encoder, embedding databases to store the text embeddings extracted from the LLM encoder, and a classifier to map the text embedding to a classification decision.}
    \label{fig:system-architecture}
\end{figure*}

To effectively discern machine-generated from human-generated text, we leverage the power and sophistication of Large Language Model (LLM) encoders. These encoders, which have been proven adept at generating dense and semantically rich embeddings of textual data \cite{wijesiriwardene2023analogical}, serve as the foundation of our methodology. By feeding both human and machine-generated texts into such an encoder, we extract embeddings that inherently capture the unique characteristics and intricacies of each text type. These embeddings are then stored for further analysis. Atop these embeddings, we construct classifiers from Multilayer Perceptrons (MLPs) and K-Nearest Neighbors (KNN) to more advanced, contrastive learning-based KNN classifiers. We call this approach T5LLMCipher, based on the T5 \cite{raffel2020exploring} LLM encoder that we use in our implementation. In the ensuing sections, we describe LLM encoders, elucidate why they stand out as promising tools for machine-generated text detection, and provide a detailed exposition on the classifiers designed atop these embeddings. The overall proposed system architecture is shown in Figure~\ref{fig:system-architecture}.


\subsection{LLM Text Encoder}


Modern language models have shown impressive capabilities in various tasks, ranging from text generation to translation and summarization ~\cite{touvron2023llama}. The strength of such models lies not just in their ability to generate coherent text, but also in their capacity to comprehend and encode the semantic essence of given input text into dense representations, known as embeddings \cite{neelakantan2022text}. An encoder in an LLM serves as a crucial component that transforms raw text input into these high-dimensional embeddings. Such embeddings capture intricate patterns, contextual relationships, and even abstract concepts from the text, making them a rich source of information. As depicted in Equation~\ref{eq:encoder} the text encoder $f$  transforms a text sample $x$ into a representation $e$ , which encapsulates the semantic essence of the input.

For our objective of building a system to distinguish machine-generated text, leveraging the power of these encoders offers a strategic advantage. As we aim to distinguish between human-generated and machine-generated content, it's imperative to obtain representations that capture the nuanced differences between these two types of content. Given the massive amount of data and diverse linguistic structures these LLMs are trained on, their encoders are likely to offer embeddings that can discern subtle differences in text structure and coherence \cite{hwang2023embedtextnet}. Furthermore, using a frozen encoder from an established LLM provides a standardized way to obtain these embeddings without the need to train our own encoding mechanism. This not only reduces the computational burden but also ensures that we leverage state-of-the-art representations. For these reasons, in our approach, we employ the encoder from an LLM with an encoder-decoder architecture. The choice to avoid purely decoder-based LLMs is deliberate, as we aim to directly process and embed a given text sample without the need for generation.

Let $LLMEncoder(\cdot)$ denote a frozen LLM text encoder, which maps an input text sample $x$ to an embedding space:
\begin{equation}
    \label{eq:encoder}
    e = LLMEncoder(x)
\end{equation}


\subsection{Human vs. LLM Classifiers}

Building on the rich embeddings provided by the LLM encoders, our methodology uses a range of classifiers, each designed to utilize the distinct characteristics of these embeddings. The following subsubsections outline the different types of Cipher models that we use to build off of these embeddings.

\subsubsection{Multilayer Perceptron Classification}

At the heart of our classification methodology lies the embeddings extracted from LLM encoders, dense representations with information necessary to differentiate between human and machine-generated texts. Multilayer Perceptrons (MLPs)~\cite{lecun2015deep}, with their multiple layers of interconnected neurons, serve as our one of our tools to navigate and interpret these embeddings. While the embeddings capture the core distinctions between text types, the MLPs facilitate the transformation of this latent information into actionable classification decisions. They do this by modeling intricate relationships present in the embeddings and leveraging their inherent capacity to extract and amplify higher-level features, effectively bridging the gap between raw embeddings and the final classification outcome. The flexibility and adaptability of MLP architectures, coupled with their ability to generalize over complex datasets, position them as a natural choice for our initial classification endeavors on the embeddings.

\subsubsection{K-Nearest Neighbors Classification}

We also use the K-Nearest Neighbors (KNN) classifier to distinguish between human and machine-generated texts. At its core, the KNN algorithm attributes a classification label to an unseen sample based on the predominant label of its  $k$ closest neighbors. One of the innate strengths of the KNN algorithm is its resistance to overfitting, an issue that often plagues more complex classifiers. Models that are highly susceptible to overfitting excel on their training data but display a marked decline in performance on novel datasets \cite{abouelnaga2016cifar}. Several advanced classifiers, despite their sophistication, are known to fall into this trap \cite{bejani2021systematic}. The non-parametric nature of KNN circumvents this challenge by drawing on broader patterns and trends present in the data. We carefully choose the training data for our models to be paired in nature, where every human-generated text is coupled with a corresponding machine-generated text on the same topic. In the training datasets that are used for our models, each sample is a pair consisting of human-written text and the machine-generated text that was generated using this human-written text. This structure is particularly beneficial for KNN. When faced with paired texts that share thematic content but differ in their origins, KNN can sharply focus on the subtle distinctions between them. During inference, the classification decision should predominantly reflect the origin of the text rather than its content.

\subsubsection{Contrastive K-Nearest Neighbors}
We also employ a contrastive learning strategy, a technique rooted in the idea of understanding and distinguishing between similar and dissimilar data representations. The essence of contrastive learning is to pull together the representations of similar data while pushing apart representations of dissimilar data ~\cite{radford2021learning}. In our context, we leverage this methodology to explicitly accentuate the differences between human and machine embeddings. The rationale behind this is twofold: First, by emphasizing these differences, we can create a clearer distinction in the latent space between human and machine-generated content. Second, this distinct separation facilitates the KNN classifier's decision-making process, as it can more confidently assign labels based on proximity in this differentiated space. As a result, we aim to achieve enhanced classification accuracy and robustness against potential ambiguities between human and machine data points.

On top of the LLM encoder \( LLMEncoder(\cdot) \), we add a series of fully connected layers, collectively represented as \( g(\cdot) \), to process the embeddings further. Let \( h = g(e) \) be the output after all these fully connected layers. The objective of these layers is to refine the embeddings such that they are more discriminative for our task.

Let the final layer's output of \( h \) be represented as \( z \). This vector \( z \) serves as the refined embedding to be used for classification purposes.

Given a pair of samples $(x_i, x_j)$, we first map them to their respective embeddings:
\begin{align}
    e_i &= LLMEncoder(x_i) \\
    e_j &= LLMEncoder(x_j)
\end{align}

The Euclidean distance between two embeddings is given by:
\begin{equation}
    D(e_i, e_j) = \|e_i - e_j\|_2
\end{equation}

We then construct a similarity matrix $M$ where each element $M_{ij}$ is given by:
\begin{equation}
M_{ij} = 
\begin{cases} 
    1 & \begin{array}{l}
        \text{if both } x_i \text{ and } x_j \text{ are human-generated }\\
        \text{or both are machine-generated}
        \end{array} \\
    0 & \begin{array}{l}
        \text{if one of } x_i \text{ and } x_j \text{ is human-generated}\\
        \text{and the other is machine-generated}
        \end{array}
\end{cases}
\end{equation}

The Triplet Margin Loss for a given anchor \( v_a \), positive sample \( v_p \), and negative sample \( v_n \) is defined as:
\begin{equation}
    \mathcal{L}_{triplet}(v_a, v_p, v_n) = \max(0, D(v_a, v_p) - D(v_a, v_n) + m)
\end{equation}
Where the $m$ is the margin hyperparameter, and \( D \) is the function that computes the Euclidean distance between embeddings. The goal during training is to minimize this loss, ensuring that positive pairs are closer than negative pairs in the embedded space by at least a predefined margin.

With the embeddings \( z \) obtained, we employ a K-Nearest Neighbors (KNN) classifier to distinguish between human and machine-generated texts.

The Euclidean distance between these contrastively learned embeddings is used for the KNN classification:
\begin{equation}
    D(z_i, z_j) = \|z_i - z_j\|_2
\end{equation}
Where \( z_i \) and \( z_j \) are the embeddings corresponding to text samples \( x_i \) and \( x_j \), respectively.

In high-dimensional spaces, understanding the differences between embeddings can be approached from two primary perspectives: their orientation (direction) and their relative position (magnitude). Cosine similarity, and its derivative, cosine distance, focuses primarily on the angle between two vectors, effectively ignoring their magnitude. While this can be useful in scenarios where the magnitude (or the length) of the vectors is not informative or might introduce noise, it can miss nuances in situations where the magnitude contains discriminative information. On the other hand, Euclidean distance offers a more comprehensive measure as it accounts for both the orientation and magnitude of vectors in the high-dimensional space. By doing so, it captures the full spectrum of relationships between vectors, making it sensitive to more subtle variations and characteristics. 
Therefore, using Euclidean distance in our contrastive KNN model provides a more robust and encompassing evaluation of the distinctions between the text embeddings, enabling a more effective differentiation between human and machine-generated text.

\section{System Implementation and Evaluation}
\label{experiments}

\subsection{Implementation}

We utilized the Google Flan-T5-XL implementation available on Huggingface~\cite{flan-t5-xl} to generate text embeddings. In our "in-the-wild" experiment detailed in Section~\ref{sec:in-the-wild}, we also examined the impact of replacing the T5 text embedder by switching to RoBERTa-base~\cite{roberta-base}. T5 employs a SentencePiece tokenizer~\cite{kudo2018sentencepiece}, which processes inputs as a continuous stream, obviating the need for prior tokenization. In contrast, RoBERTa leverages byte-level Byte-Pair Encoding (BPE)~\cite{sennrich2016neural} for its tokenization. BPE, initially conceived to accommodate a vast range of vocabularies, focuses on sequentially merging the most frequent character pairs to create new tokens. Both T5 and RoBERTa implementations restrict their inputs to a maximum size of 512 tokens. To extract a text embedding from T5, we target the last hidden state in the encoder. In transformer models such as T5~\cite{raffel2020exploring}, each layer successively captures increasingly abstract and high-level features. By the time data reaches the last hidden state, the model has assimilated a wealth of contextual nuances from both micro and macro perspectives, representing the entire text sequence. For RoBERTa, we obtain the embedding from the hidden state of the last layer's first token (CLS token). This particular hidden state is widely recognized for its ability to represent a given text and is thought to encapsulate the overarching semantic information of the input sequence~\cite{tan2022domain, zou2021unsupervised, hong2022lea}. The resulting T5 embedding yields a vector of size 2048, while the RoBERTa embedding produces a vector of size 768.

For our MLP, we utilize a fully connected feedforward network, consisting of six layers. The network architecture funnels down from an input size of 2048 to an output size of 6 for the 6-class multiclass classifier, an output size of 5 for the 5-class multiclass classifier, or an output size of 2 for the binary classifier. Each layer, except for the last, employs the ReLU activation function. For training, the model uses Cross Entropy Loss as its loss function, optimized using the Adam algorithm with a learning rate of 0.0001, all implemented using Pytorch~\cite{paszke2019pytorch}. All of the MLP models are trained and tested directly on the LLM embeddings without any additional processing. The datasets for the MLP models are partitioned into training, validation, and testing subsets. Specifically, 80\% is allocated for training where the model is trained for 500 epochs, 10\% serves as a validation set where the model demonstrating the highest validation accuracy is saved, and the remaining 10\% is reserved for testing. We refer to the multiclass MLP classifiers as T5LLMCipher-MC and the binary MLP classifiers as T5LLMCipher-Bi.

For our KNN experiments, we implement it with the scikit-learn~\cite{scikit-learn} library, using K = 5 with the Euclidean distance function. For the contrastive KNN We employed a straightforward MLP model comprising 150M parameters, using ReLU activations between layers. The output layer, of size 512, had no activation. The TripletMarginLoss was our chosen loss function. In our training (80\%) and validation (10\%) splits, each embedding served as an anchor once per epoch. Positives were randomly selected from the same class as the anchor, while negatives were from different classes. The model was trained for 100 epochs using the Adam optimizer~\cite{kingma2014adam} with a learning rate of 0.0001. We refer to the standard KNN trained on just the T5 embeddings as T5LLMCipher-KNN and the contrastive KNN as T5LLMCipher-C-KNN.

\subsection{Evaluation}

We conduct a series of comprehensive experiments designed to evaluate the performance and robustness of our proposed approach under various scenarios. Initially, an experiment evaluates the potential to attribute text generations to known models. Next, a cross-domain experiment assesses the model's capability to generalize and adapt across different text domains. Following this, we perform a cross-generator experiment to investigate the model's effectiveness in distinguishing text originating from text generation models. An in-the-wild data experiment is also conducted to simulate real-world conditions where the data is from unknown generators and domains, testing the model's robustness and reliability in practical applications. To further probe the model's resilience, an adversarial robustness experiment is carried out, exposing the model to intentionally crafted adversarial inputs to explore our method's robustness. Lastly, we present a t-SNE visualization of our multiclass MLP classifier to show that it effectively separates human and machine texts on known data.

\subsection{Text Generator Identification}
\begin{table}[t!]
\centering
\resizebox{\columnwidth}{!}{
\begin{tabular}{r|rrrrrr}
\specialrule{1pt}{0pt}{-1pt}
& Bloomz & ChatGPT & Cohere & Davinci & Dolly & Human \\
\specialrule{1pt}{0pt}{0pt}
Bloomz  & 1238   & 0       & 3      & 8       & 15    & 8     \\
ChatGPT & 0      & 1068    & 55     & 110     & 13    & 1     \\
Cohere  & 3      & 11      & 940    & 15      & 57    & 8     \\
Davinci & 1      & 109     & 19     & 1039    & 64    & 29    \\
Dolly   & 19     & 14      & 91     & 49      & 978   & 74    \\
Human   & 0      & 0       & 3      & 5       & 28    & 6620  \\
\specialrule{1pt}{0pt}{-1pt}
\end{tabular}
}
\caption{Confusion matrix for identifying the source generator using the T5LLMCipher-MC classifier}
\label{tab:confusion_matrix}
\end{table}

\begin{table}[b!]
\centering
\resizebox{\columnwidth}{!}{
\begin{tabular}{r|rrrr}
\specialrule{1pt}{0pt}{-1pt}
& Precision & Recall & F1-Score & Support \\
\specialrule{1pt}{0pt}{0pt}
Bloomz  & 98.2      & 97.3   & 97.7     & 1272    \\
ChatGPT & 88.9      & 85.6   & 87.2     & 1247    \\
Cohere  & 84.6      & 90.9   & 87.6     & 1034    \\
Davinci & 84.7      & 82.4   & 83.6     & 1261    \\
Dolly   & 84.7      & 79.8   & 82.2     & 1225    \\
Human   & 98.2      & 99.5   & 98.8     & 6656    \\
\specialrule{1pt}{0pt}{-1pt}
\end{tabular}
}
\caption{Results with respect to each class for identifying the source generator}
\label{tab:attribution}
\end{table}

We train a 6 class T5LLMCipher-MC classifier to distinguish between each generator, trained on M4 data across the Arxiv, PeerRead, Reddit, Wikihow, and Wikipedia domains.

Table~\ref{tab:confusion_matrix} is a confusion matrix that details the number of times a given text source was classified as each possible source. The diagonal of the matrix represents true positives or instances where the classifier correctly identified the source. For example, the classifier correctly identified Bloomz-generated text 98.1\% of the time. Human-generated text also seems to be distinctly characterized, as the classifier correctly attributed human texts to being human-generated 98.2\% of the time.

\begin{figure*}[]
    \centering
    \includegraphics[width=1\linewidth]{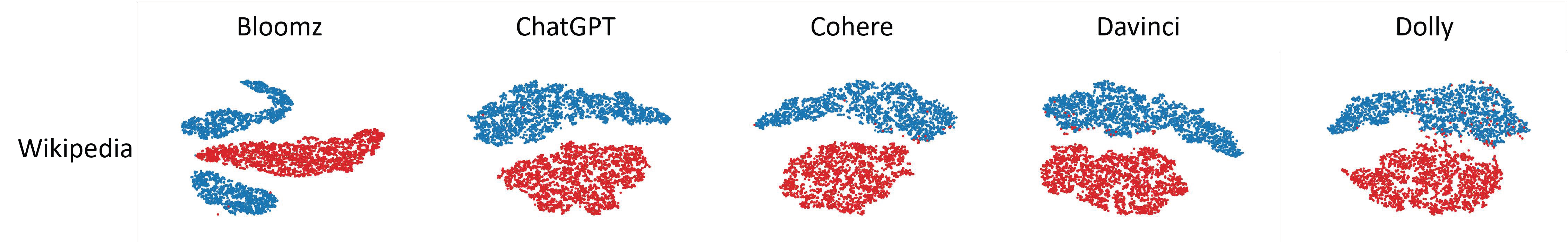}
    \caption{t-SNE visualization of T5LLMCipher-MC classifier embeddings illustrating the distribution of Human (blue), and Machine (red) generated text. There exists a clear separation between Human and Machine generated text.}
    \label{fig:6_class_embeddings_1}
\end{figure*}

\begin{figure*}[]
    \centering
    \includegraphics[width=1\linewidth]{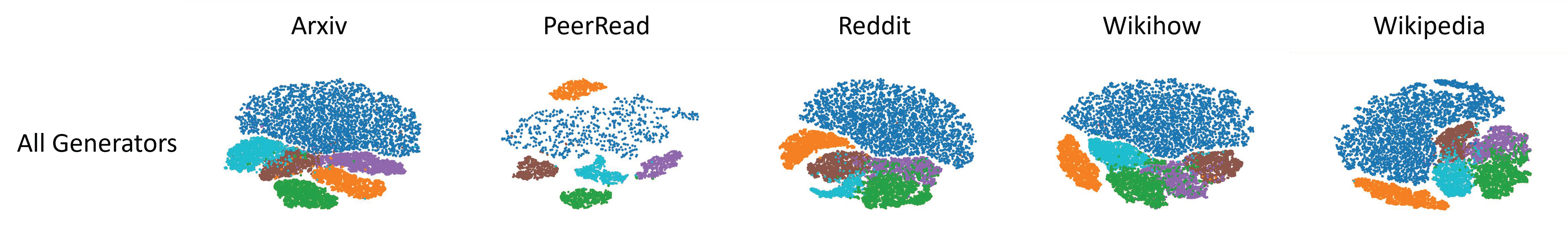}
    \caption{t-SNE visualization of T5LLMCipher-MC classifier embeddings illustrating the distribution of Human (blue), Bloomz (orange), ChatGPT (green), Cohere (teal), Davinci (purple), and Dolly (brown) generated texts across different domains. Our T5LLMCipher-MC classifier shows a strong ability to distinguish the source generator of text.}
    \label{fig:6_class_embeddings_2}
\end{figure*}

Table~\ref{tab:attribution} delves into the individual performance metrics of the classifier for each text generator. It lists precision, recall, F1-Score, and support for each class. This table shows that the classifier performs exceptionally well in identifying Human and Bloomz text, with F1 scores of 0.988 and 0.977, respectively. ChatGPT and Cohere texts are fairly distinguishable with F1-Scores above 0.870, whereas Davinci and Dolly have the lowest F1-Scores, suggesting these might be harder to distinguish or have characteristics that overlap with other classes. Additionally, we visualize the model's proficiency in differentiating between machine-generated and human-generated texts through the t-SNE presented in Figure~\ref{fig:6_class_embeddings_1}. For this visualization, we extracted features from the layer before the classification layer, which has a vector size of 256. The domains showcased include Arxiv and Wikihow, spanning five distinct generators. In this figure, human and machine samples manifest as more defined clusters than the distribution observed in Figure~\ref{fig:t-SNE_motivation_1}. We further visualize the model's ability to attribute text to source generators in each domain in Figure~\ref{fig:6_class_embeddings_2}. There we show the t-SNE visualization of 6 class classifier embeddings on each of the different generator's texts and see that each one forms a distinct cluster with minimal overlap. Compared to the original embedding space shown in Figure~\ref{fig:t-SNE_motivation_2}, we now see excellent separation between classes. Overall, these experiments show that the identification of the source generator of texts is feasible, and that different generators have unique identifiable characteristics, or "fingerprints," within their generated texts.

\subsection{Cross-Domain Experiment}

\label{sec:cross_domain}
\begin{table}[b!]
\centering
\resizebox{\columnwidth}{!}{
\begin{tabular}{r|rrrrr}
\specialrule{1pt}{0pt}{-1pt}
                       & Arxiv & PeerRead & Reddit & Wikihow & Wikipedia \\
\specialrule{1pt}{0pt}{0pt}
T5LLMCipher-MC & 73.3  & 86.0     & 85.6   & 70.9    & 72.7      \\
T5LLMCipher-Bi & 75.7  & 83.8     & 86.4   & 75.0    & 71.4      \\
T5LLMCipher-C-KNN        & 75.4  & 87.0     & 90.0   & 76.4    & 74.2      \\
T5LLMCipher-KNN                    & 64.9  & 72.0     & 75.7   & 54.2    & 68.2      \\
\specialrule{1pt}{0pt}{-1pt}
\end{tabular}
}
\caption{Cross-domain experiment, showing the F1 score for the machine-generated text class where models are trained on four domains and tested on the domain it wasn't trained on. Each column shows the domain that is being tested on.}
\label{tab:cross_domain}
\end{table}

In our cross-domain experiment, as illustrated in Table~\ref{tab:cross_domain}, we evaluated the capability of our classifiers to generalize in identifying machine-generated text from unseen domains. These classifiers were trained and tested using samples from the Human class and the five machine text generators: ChatGPT, davinci-003, Cohere, Dolly, and BLOOMz. The 6 class T5LLMCipher-MC classifier was trained to attribute text to the Human class and each of the five machine-text generators. For the testing of the 6 class T5LLMCipher-MC classifier in the cross-domain experiment, if the classifier detects a sample as a human sample, it is treated as a human sample, and if it is detected as any of the five machine generators, it is treated as a machine sample. All other classifiers were trained and tested on the binary classification task of distinguishing between machine and human text.

In this experiment, the detectors were trained on four of the domains, and evaluated to detect machine-generated text in the domain that the detector wasn't trained on. The results on the 6-Class T5LLMCipher-MC classifier, T5LLMCipher-MC classifier, and the T5LLMCipher-C-KNN classifier demonstrated closely matched performance across multiple domains, suggesting a robustness in their adaptability. The generalization to unseen domains appears to vary between the test datasets, with Wikihow appearing to be the most difficult to generalize to. In this experiment, the T5LLMCipher-C-KNN classifier performed the best, with an average F1-score of 80.6 to unseen domains.

\begin{table}[h!]
\centering
\resizebox{\columnwidth}{!}{
\begin{tabular}{r|rrrrr}
\specialrule{1pt}{0pt}{-1pt}
                       & Bloomz & ChatGPT & Cohere & Davinci & Dolly \\
\specialrule{1pt}{0pt}{0pt}
T5LLMCipher-MC & 76.0  & 96.3      & 92.6  & 84.4   & 69.8  \\
T5LLMCipher-Bi & 79.7  & 97.0      & 92.0  & 85.3   & 77.8  \\
T5LLMCipher-C-KNN        & 78.4  & 99.4   & 92.5  & 85.8   & 78.1 \\
T5LLMCipher-KNN                    & 59.2     & 87.6      & 82.1     & 72.7      & 68.7    \\
\specialrule{1pt}{0pt}{-1pt}
\end{tabular}
}
\caption{Cross-generator experiment, showing the F1 score with respect to the machine-generated text class where models are trained on four generators and tested on the generator it wasn't trained on. Each column shows the generator that is being tested on.}
\label{tab:cross_generator}
\end{table}
\subsection{Cross-Generator Experiment}

In our cross-generator experiment, as illustrated in Table~\ref{tab:cross_generator}, we evaluated the capability of our classifiers to generalize in identifying machine-generated text from unseen generators. These classifiers were trained and tested using samples from all of the domains, which were Wikipedia, Reddit, WikiHow, PeerRead, and Arxiv. The 5 class T5LLMCipher-MC classifier was trained to attribute text to Human class and four of the machine generators and then tested on the one generator it was not trained on. For the testing of the 5 class T5LLMCipher-MC classifier in the cross-domain experiment, if the classifier detects a sample as a human sample, it is treated as a human sample, and if the machine sample that it wasn't trained on is detected as any of the four machine generators it was trained on, it is treated as a machine sample. All other classifiers were trained and tested on the binary classification task of distinguishing between machine and human text. In this experiment, Bloomz and Dolly appear to be the most difficult generators to generalize to, while ChatGPT looks to be the easiest to generalize to. Once again, in this experiment we find that the T5LLMCipher-C-KNN classifier narrowly beats out the other classifiers in generalizing to unseen generators, with an average F1 score of 86.8.

\subsection{In-the-wild Experiment}
\label{sec:in-the-wild}


\begin{table}[b!]
\centering
\resizebox{\columnwidth}{!}{
\begin{tabular}{r|rrrr|r}
\specialrule{1pt}{0pt}{-1pt}
                         & AI Writer & Article Forge & Kafkai & Reddit Bot & Avg \\
\specialrule{1pt}{0pt}{0pt}
T5LLMCipher-MC              & 62.3     & 84.4         & 61.8  & 84.6      & \textbf{73.3} \\
T5LLMCipher-Bi              & 59.2     & 78.8         & 46.2  & 82.1      & 66.6 \\
T5LLMCipher-C-KNN          & 57.2     & 83.2         & 49.2  & 82.9      & 68.1 \\
T5LLMCipher-KNN                      & 37.2     & 45.4         & 32.8  & 67.0      & 45.6 \\
RoBERTaLLMCipher-MC & 61.5  & 81.2            & 41.9     & 81.8         & 66.6 \\
\specialrule{1pt}{0pt}{0pt}
BERT-D                       & 68.6        & 70.6              & 55.6       & 67.3         & 65.5 \\
GLTR-BERT                  & 49.2        & 83.8              & 29.0       & 73.8         & 59.0 \\
GLTR-GPT2                 & 29.6        & 83.1              & 29.2       & 83.5         & 56.4 \\
RoBERTa-D                & 20.9        & 48.7              & 37.6       & 76.8         & 46.0 \\
RADAR                       & 50.2        & 61.6              & 60.2       & 18.0         & 48.0 \\
Fast-DetectGPT            & 32.1        & 52.9              & 98.7       & 97.5         & 70.3 \\
\specialrule{1pt}{0pt}{-1pt}
\end{tabular}
}
\caption{Experiment comparing F1 score with respect to machine-generated text of existing state-of-the art approaches to our method on in-the-wild text.}
\label{tab:in_the_wild}
\end{table}

To demonstrate our approach on in-the-wild data, we use the in-the-wild machine-generated text data released by Pu et al.~\cite{pu2023deepfake}. Overall, three of these datasets consist of news article styled articles and one of the datasets consists of social media styled posts. The machine-generated text from this dataset were generated from API and while it is unknown what the exact model architecture was of these generations, the ArticleForge and Kafkai models were indicated to have used variants of GPT-2 models. Details of this dataset are discussed in Section~\ref{Data}. Given the available information, there is not known to be an overlap in domain or generators with the M4 dataset~\cite{wang2023m4}. While both datasets contain data sourced from Reddit, the Reddit ELI5 data from M4 contains machine-generated text in the form of answers to questions, while the RedditBot data from Pu et al.~\cite{pu2023deepfake} contains machine-generated text in the form of social media posts.

We train our models on the five generators across five domains from the M4 dataset. The 6 class T5LLMCipher-MC was trained to attribute text to Human class and each of the five machine-text generators. This model, which was trained to attribute texts to known generators, will instead be used for the binary task of attributing texts as machine-generated or human-generated. For the testing of the 6 class T5LLMCipher-MC classifier in the cross-domain experiment, if the classifier detects a sample as a human sample, it is treated as a human sample, and if it is detected as any of the five machine generators, it is treated as a machine sample. The rationale behind this testing strategy for our T5LLMCipher-MC is that these new samples from unseen LLMs will likely be far away from the human text cluster and may share some characteristics with one of many known generators that are already included in its training. All other classifiers were trained and tested on the binary classification task of distinguishing between machine and human text. Furthermore, to demonstrate the impact of the backbone LLM encoder, we also test a 6 class MLP classifier using RoBERTa instead of T5, which is indicated as RoBERTaLLMCipher-MC. We also compare against the four models discussed in Section \ref{motivation} (BERT-D, GLTR-BERT, GLTR-GPT2, and RoBERTa-D) \cite{pu2023deepfake} by training these models on the same training dataset used to train our classifiers to provide an apples-to-apples comparison against existing supervised detection models. To compare against black-box detection methods, we examine the RADAR model \cite{hu2023radar}, a framework designed for detecting machine-generated texts through adversarial learning. Central to RADAR are two key components: a paraphraser and a detector. The paraphraser, built on the T5-large model \cite{raffel2020exploring}, is tasked with altering machine-generated text in an effort to bypass detection. Concurrently, the detector, leveraging the RoBERTa-large model \cite{liu2019roberta}, is trained to identify both the original AI texts and those that have been paraphrased. This adversarial training loop ensures that RADAR continuously evolves in its ability to discern AI-generated content. The training data for RADAR comprises a human text corpus, sourced from the OpenWebText dataset \cite{OpenWebTextCorpus2019}, and machine-generated text, which employs the Vicuna-7B model to create texts based on prefixes extracted from the human-text corpus.  We further compare against Fast-DetectGPT \cite{bao2023fast}, another black-box machine-text detection algorithm. This method quantifies the curvature of the conditional probability function, which is indicative of the likelihood of text being machine-generated based on the comparative evaluation of the probabilities of given and alternative token choices. The positive curvature at a point suggests a higher likelihood of the text being machine-generated, whereas a close-to-zero curvature indicates human authorship. GPT-Neo 2.7B \cite{black2022gpt} was used as the scorer model to obtain token probabilities.

We find that T5LLMCipher-MC provides the best results with an average F1 score of 73.3\% across the four in-the-wild datasets. We also observe that our models have the toughest time generalizing to the Kafkai generations, and the easiest time detecting RedditBot generations. We also find that T5LLMCipher-MC with T5 as the LLM encoder outperforms RoBERTaLLMCipher-MC with RoBERTa as the LLM encoder. Some other notable results come from Fast-DetectGPT, where Kafkai and Reddit Bot generations are detected with very high F1 scores. This can be explained by the underlying scoring model used by Fast-DetectGPT, which is a GPT variant. Kafkai and Reddit Bot machine-texts use models from the GPT family. This corroborates with the results from the work of Bao et al. \cite{bao2023fast} where they show that their detection methodology generalizes well to other machine text generators from the same family. However, their method is shown to struggle in discerning human from machine-generated texts from different families of LLMs, as is shown from the poor detection performance on AI Writer and Article Forge texts.

\subsection{Adversarial Robustness Experiment}

\begin{table}[b!]
\centering
\resizebox{\columnwidth}{!}{
\begin{tabular}{r|rrrr|r}
\specialrule{1pt}{0pt}{-1pt}
                & AI Writer & Article Forge & Kafkai & Reddit Bot & Avg \\
\specialrule{1pt}{0pt}{0pt}
T5LLMCipher-MC     & 54.9       & 77.3          & 54.9   & 81.5       & \textbf{67.2} \\
T5LLMCipher-Bi     & 62.4       & 66.9          & 22.8   & 50.0       & 50.5 \\
T5LLMCipher-C-KNN  & 31.0       & 53.0           & 22.4   & 53.1       & 39.9 \\
T5LLMCipher-KNN    & 19.8       & 23.7           & 37.3   & 37.3       & 29.5 \\
\specialrule{1pt}{0pt}{0pt} 
BERT-D             & 54.6      & 55.2          & 52.1   & 58.8       & 55.2 \\
GLTR-BERT          & 29.7      & 65.9          & 10.9   & 73.0       & 44.9 \\
GLTR-GPT2          & 9.1      & 57.3          & 10.8   & 76.9       & 38.5 \\
RoBERTa-D          & 13.0      & 36.6          & 29.1   & 82.6        & 40.3 \\
RADAR              & 49.6       & 64.3           & 58.9    & 36.2      & 52.3 \\
Fast-DetectGPT     & 29.1      & 54.0          & 54.0   & 97.7        & 58.7 \\
\specialrule{1pt}{0pt}{-1pt}
\end{tabular}
}
\caption{Experiment comparing F1 score with respect to machine-generated text of existing state-of-the art approaches to our method on adversarially perturbed in-the-wild text.}
\label{tab:adv_robustness}
\end{table}

To demonstrate our method's robustness to adversarial attacks, we evaluate our system against a recently introduced perturbation attack method for evading machine-generated text detection, DFTFooler~\cite{pu2023deepfake}. DFTFooler carefully modifies machine-generated text to evade detection while attempting to preserve linguistic quality. This is achieved by pinpointing words in the text which are confidently predicted by a pre-trained LLM and substituting them with synonyms that are semantically similar yet less confidently predicted. The synonym selection process adheres to a meticulous protocol to uphold linguistic quality: First, potential synonyms are extracted based on cosine similarity between word embeddings, ensuring a semantic alignment with the original word. Subsequent steps involve part-of-speech checking, to maintain grammatical congruity, and semantic similarity checking to ensure the substituted word does not deviate semantically from the original text. The selected synonym is then one that is predicted with lower confidence by the language model, which alters the predictability of the word choices in the text. Notably, unlike other adversarial techniques, DFTFooler achieves this without necessitating any queries to the detection model, which makes it an attractive choice for potential adversaries who would want to attack a black-box detection model~\cite{gao2018black}. We use machine-generated perturbed texts using DFTFooler, setting the low probability threshold to 0.01, the synonym similarity threshold to 0.7, and allowing for 10 word perturbations. We show the results of this attack on the same datasets of in-the-wild data as in Section~\ref{sec:in-the-wild} on the  T5LLMCipher-MC, T5LLMCipher-Bi, T5LLMCipher-C-KNN and T5LLMCipher-KNN classifiers, as well as test this attack on the other state-of-the art classifiers tested in Section~\ref{sec:in-the-wild}. We randomly sample 200 texts from the human-generated texts and 200 texts from the machine-generated texts from each of the four datasets for a total of 1600 samples. We show the results of this experiment in Table~\ref{tab:adv_robustness}, where we report the F1 score with respect to the machine-generated text class. These results show that the T5LLMCipher-MC classifier was the most adversarially robust model, with an F1 score of 67.2. This represents a reduction of just 6.1 points from the non-adversarially perturbed in-the-wild text from the experiment in Table~\ref{tab:in_the_wild}.

\section{Discussion and Future Work}
\label{discussion}

Interestingly, we found that the T5LLMCipher-C-KNN outperformed the T5LLMCipher-KNN, T5LLMCipher-MC and the T5LLMCipher-Bi classifiers on the cross-domain and cross-generator experiments. The performance of the T5LLMCipher-C-KNN can be rationalized by the theoretical underpinning of its design, where the primary goal was to capture the inherent differences between human and machine-generated texts while being agnostic to their specific domains or the machine generators behind them. By leveraging a contrastive learning paradigm, the system was tailored to magnify the subtle nuances between these two categories. Contrastive learning inherently emphasizes on pulling apart representations of dissimilar pairs while bringing closer the representations of similar pairs. In the context of our application, this means that human-generated text embeddings are encouraged to be distinct from any machine-generated text, regardless of the particular machine generator or domain from which it originated. On the other hand, all human texts, irrespective of their domain, would be represented closely in the embedding space, and similarly for machine texts from various generators. When transitioning to cross-domain and cross-generator scenarios, this foundational design proved to be beneficial. Instead of being overly attuned to the specific characteristics of the training domain or generator, the T5LLMCipher-C-KNN was focused on the broader distinction between human and machine. Consequently, it was more resilient to shifts in domain or the introduction of texts from unfamiliar machine generators.

However, even though the T5LLMCipher-MC classifier fell short of the T5LLMCipher-C-KNN classifier by a small margin in both the cross-domain and cross-generator experiments, it provided the best detection performance in both the in-the-wild and adversarial robustness experiments. The in-the-wild experiment can be considered a joint case of a cross-generator and cross-domain experiment because the generator and domains are both unknown to the detectors, so it was initially surprising that the model that performed the best on the cross-generator and cross-domain experiment did not perform the best on the in-the-wild experiment. Even though the T5LLMCipher-MC classifier was broadly categorizing texts as human or machine during testing, the knowledge acquired from distinguishing among multiple machine generators added depth to its understanding. Hence, it was better equipped to generalize and accurately classify texts, even when the specifics of their origin were unknown or when the texts were adversarially perturbed. This finer-grained knowledge may have allowed the T5LLMCipher-MC classifier to outperform its binary counterparts in these particular settings. In simpler terms, each machine generator introduces unique elements into the content's semantics. When we group machine-generated texts separately by generator in the latent space, it becomes more straightforward to distinguish between human-generated and machine-generated texts. Additionally, this approach aids in generalizing to unseen generators, enhancing our ability to discern their origin using the sub-clustering of known machine-generated text artifacts rather than treating all machine-generators as a single cluster.


The results of the text generator attribution experiment in Table~\ref{tab:attribution} and Table~\ref{tab:confusion_matrix} show that it is possible to not only distinguish between human and machine-generated text, but it is also possible to attribute the text to a specific generator. Our framework successfully attributes authorship to specific generators included in the T5LLMCipher training data. However, for content produced by machine text generators not included in the initial training, while our system can recognize the content as not human-generated, it cannot pinpoint the exact generator. Detecting a novel generator necessitates retraining of T5LLMCipher for accurate attribution. Fortunately, this process is computationally lightweight. 

In this study, our primary focus was to emphasize the efficacy of LLM encoder embeddings in detecting machine-generated text, rather than on the complexity of the classifiers. 
Notably, our results show state-of-the-art generalization and robustness against adversarial attacks, despite the relatively simple nature of the MLP and KNN classifiers. This underscores the inherent quality and potential of the embeddings themselves towards the task of detecting machine-generated text. Given these promising outcomes with simple models, future research that explores more sophisticated architectures in tandem with the LLM encoder embeddings holds significant potential to push generalizability and robustness even further. Additionally, adversarial training strategies could potentially be utilized to improve adversarial robustness of classifiers built on top of these embeddings.

\subsection{Ethical Considerations}

The deployment of human versus machine text detection systems warrants important ethical considerations when deployed, particularly in light of the potential for false positives where genuine human-written texts are mistakenly identified as machine-generated. Such misclassifications carry significant ramifications, including but not limited to, the erosion of trust in automated systems, potential harm to individuals' reputations, and the broader societal impact of inadvertently censoring legitimate expressions in digital communications. In educational contexts, misattributed authorship due to AI-generated content can lead to significant consequences, as seen in an incident where a professor threatened to fail his entire class based on a machine-text detection tool \cite{Verma2023ChatGPT}. This underscores the critical need for caution in relying on automated detectors for decision-making. Such tools are not foolproof and may lead to unjust outcomes, making it imperative to use them with caution. Machine-text detection tools have also been shown to be biased against non-native English writers \cite{liang2023gpt}, adding even more difficulty in achieving accurate and fair evaluations of texts. In this context, it is crucial to emphasize the limitations of relying solely on systems that assess textual authenticity based on input text alone. Therefore, it is important to recognize that automated machine-text detection systems should not be used for autonomous decision making without human supervision, especially in high-stakes scenarios. 


\section{Conclusion}
\label{conclusion}
The ever-growing prevalence of LLMs has sparked significant interest in developing reliable tools to detect machine-generated text. This necessity is accentuated by the realization that many existing detectors, although promising, stumble when faced with real-world complexities. Specifically, their capability to generalize over text from a multitude of generators and spanning various domains is called into question. Addressing this challenge has been central to our work, where we undertook a comprehensive exploration of machine-generated text detection across varying generators and domains. Our observations, primarily from the distribution patterns seen in t-SNE visualizations of embeddings from a pretrained LLM's encoder, signaled a promising direction. These patterns indicated the potential of these embeddings to differentiate between human-authored and machine-crafted texts. Acting on this observation, we proposed a novel system for the detection of machine-generated content, harnessing the encoder of a pretrained LLM. Our extensive evaluations, spanning nine machine-generated text systems and an equal number of domains, underscored the robustness of our approach where we show state-of-the-art generalization performance to detecting machine-generated text.

\section*{Acknowledgments}
This research project and the preparation of this publication were funded in part by the Department of Homeland Security (DHS), United States Secret Service, National Computer Forensics Institute (NCFI) via contract number 70US0920D70090004 and by NSF Grant No. 2245983.

\bibliographystyle{plain}
\bibliography{bibliography_3}

\begin{appendices}
\section{Dataset Details}
\label{dataset_details}


\begin{table*}[]
\centering
\resizebox{0.7\textwidth}{!}{
\begin{tabular}{r|rrr}
\specialrule{1pt}{0pt}{-1pt}
Domain & Human Semantic Similarity $\uparrow$ & Human Stylistic Similarity $\uparrow$ & USE-Cosine Similarity $\uparrow$ \\
\specialrule{1pt}{0pt}{0pt}
Arxiv     & 5.52      & 5.66  & 0.67     \\
Peerread  & 3.88    & 4.54   &  0.52    \\
Reddit    & 4.53     & 4.28   & 0.65     \\
Wikihow   & 5.70     & 4.94   & 0.68    \\
Wikipedia  & 4.44   & 4.36    & 0.57     \\

\specialrule{1pt}{0pt}{-1pt}
\end{tabular}
}
\caption{The average semantic similarity and stylistic similarity score given by the human evaluations across 5 domains. "USE-Cosine Similarity" is the average cosine similarity of the pairs of human and machine texts after using the Universal Sentence Encoder (USE) to create vector representation of the texts, where a value of 0 means no semantic similarity and 1 means perfect semantic similarity.}
\label{tab:dataquality}
\end{table*}

\subsection{Prompts}
\label{sec:appendix_prompts}

The prompts given to the LLMs to create machine-generated text varied by domain. The PeerRead Reviews domain used four different prompts, to generate a set of peer reviews with different perspectives similar to a set of human generated peer reviews. Each prompt was provided to the LLM independently of the other prompts, and each prompt contains information from which to generate text similar to the human written text that the machine text is paired with. The information unique to each individual prompt is illustrated in [brackets] for illustrative purposes. Below are examples of prompts used for each domain.


\begin{itemize}
    \item \textbf{Wikipedia:} Prompts included the title of the Wikipedia article. Example Prompt: Write a Wikipedia article with the title [title], the article should at least have 250 words.
    \item \textbf{Reddit ELI5:} Prompts included the question asked in the Reddit post. Example Prompt: Write a couple of paragraphs of at least 500 characters each to address the following question. [question].
    \item \textbf{WikiHow:} Prompts included the title and headline in the Wikihow article. Example Prompt: Write a WikiHow article content given a title and a headline. Use approximately 300 words. Title: [title] Headline: [headline].
    \item \textbf{PeerRead Reviews:} Prompts included the title and abstract of the research paper. Example Prompt: Write a peer review by first describing what problem or question this paper addresses, then strengths and weaknesses, for the paper [title], its main content is as below: [abstract].
    \item \textbf{Arxiv Abstract:} Write a long abstract of a scientific paper from arXiv.org. Use approximately 200-400 words. Title: [title]. Abstract:
\end{itemize}

\subsection{Dataset Assumptions}
\label{sec:appendix_quality}

Although it is not possible to assure with absolute certainty that each entry within the human-annotated datasets was solely produced by human, it is important to note that all data were compiled before the widespread availability of sophisticated and easily accessible language models capable of producing human-like text. The most recent collection of human-written texts in the dataset we use date back to early 2022, predating the introduction of ChatGPT or Gemini.

To ensure that the machine-generated text was reasonably similar to the human-generated texts that the machine-generated texts were based on, we use the Universal Sentence Encoder (USE) ~\cite{cer2018universal} to measure cosine similarity, and human evaluations of style and semantic similarity. USE is first utilized to create high-dimensional vector representations of the texts. Then we measured the cosine similarity between the human-machine text pairs. This method is used to measure the semantic similarity of the human and machine text as an automatic evaluator. Cosine similarity values of 1 indicate a perfect semantic similarity, and values of 0 indicate no semantic similarity. In the human evaluations of text similarity, a 7-point Likert Scale test is used to measure the participant's agreement or disagreement on a scale from 1 (strongly disagree) to 7 (strongly agree) with the statements of "The two text samples are similar in style?" and "The two text samples are similar in semantics". The authors were used to evaluate the similarity of 250 human-machine text-pairs, and their scores on these statements across these text-pairs were averaged. The 250 text-pairs were randomly sampled to include 10 text-pairs across each of the combinations of five domains and five generators in the M4 dataset described in Section \ref{Data}. We report the average human evaluation scores across domains in Table \ref{tab:dataquality}. We also report the average cosine similarity across domains on the same samples used for human evaluations in Table \ref{tab:dataquality}. The average USE cosine similarity across all the evaluated samples is 0.62, which indicates that the level of semantic similarity is moderately high based on this metric. The median sentence length of the human generated and machine generated texts is 295 words and is 252 words, respectively. Given the lengths of the text samples and the expected decrease in similarity measures for longer texts generated by LLMs, the observed average similarity of 0.62 indicates a significant resemblance between the text pairs. This finding is further supported by human evaluation, as presented in Table \ref{tab:dataquality}.

As seen in Table \ref{tab:dataquality}, the average scores on stylistic and semantic similarity of the human and machine texts indicate good quality of the generated texts, with the Arxiv domain having the highest average human evaluation scores of 5.52 and 5.66 on semantic similarity and stylistic similarity, respectively. The level of similarity of texts between human and machine text pairs varied by domain, but were fairly consistent nonetheless. For instance, in PeerRead Reviews, there were bigger discrepancies between the human-generated and machine-generated samples than in other domains. One potential reason for this mismatch in semantics in the PeerRead Reviews dataset is that the machine generated peer reviews do not have access to the main body of the research papers and only consider the title and abstract of the paper, whereas the human reviewers have access to the full body of the paper in addition to the title and abstract. However, in general, the quality of the generated text seemed to be on par with the human text, even though the actual semantics did not always match on some of the datasets. We acknowledge that the evaluation of the human and machine texts in this experiment could be biased due to the author-led evaluations that might favor interpretations supporting the similarity of the texts. However, this limitation's impact on our overall system is likely minimal due to several factors. First, the USE cosine similarity measurement corroborates the author-led evaluations, providing automated confirmation of the similarity of the samples. Additionally, the model's strong performance on in-the-wild data as shown in Table \ref{tab:in_the_wild} shows that the impact of this limitation on the validity of our model's results is minimal.

\end{appendices}

\end{document}